# APC2Mesh: Bridging the gap from occluded building façades to full 3D models

Perpetual Hope Akwensi, Akshay Bharadwaj, Ruisheng Wang*

*University of Calgary, 2500 University Drive NW, Calgary, T2N 1N4, Alberta, Canada*
*\* Corresponding author. E-mail address: ruiswang@ucalgary.ca (Ruisheng Wang).*

**Abstract.** The benefits of having digital twins of urban buildings are numerous. However, a major difficulty encountered in their creation from airborne LiDAR point clouds is the effective means of accurately reconstructing significant occlusions amidst point density variations and noise. To bridge the noise/sparsity/occlusion gap and generate high fidelity 3D building models, we propose APC2Mesh which integrates point completion into a 3D reconstruction pipeline, enabling the learning of dense geometrically accurate representation of buildings. Specifically, we leveraged complete points generated from occluded ones as input to a linearized skip attention-based deformation network for 3D mesh reconstruction. In our experiments, conducted on 3 different scenes, we demonstrate that: (1) APC2Mesh delivers comparatively superior results, indicating its efficacy in handling the challenges of occluded airborne building points of diverse styles and complexities. (2) The combination of point completion with typical deep learning-based 3D point cloud reconstruction methods offers a direct and effective solution for reconstructing significantly occluded airborne building points. As such, this neural integration holds promise for advancing the creation of digital twins for urban buildings with greater accuracy and fidelity.

## 1. Introduction

The substantial surge in urban growth (DESA, 2019; Cox, 2022) has underscored the importance of digital twins (DT) – virtual models of physical assets and processes – as a highly efficient and cost-effective tool for urban management and development. Specifically, DT enables city engineers, planners, and policymakers to simulate various urban scenarios, facilitating the development of optimal solutions and informed policies to address the dynamic challenges of urban environments.

As urbanization moves in tandem with infrastructural development, DT for buildings have been explored from multiple directions, such as indoor layout, facade, roof, or full building reconstruction. For outdoor related 3D building reconstruction, point clouds (PCs) especially airborne LiDAR is generally the go-to data source. However, airborne building PCs are usually sparse with non-uniform point density, and riddled with occlusions and noise – which poses significant challenges in the accurate depiction of building surface geometry during 3D reconstruction.

To circumvent the problem of occlusion, sparsity and/or non-uniform point density, classical building reconstruction approaches (Nan and Wonka, 2017; Xie et al., 2021) usually employ primitive fitting algorithms while typical deep learning (DL) -based approaches resolve it via point cloud completion (PCC) networks (Yuan et al., 2018; Pan et al., 2021; Cai et al., 2022). However, most PCC models are built with input point sets that are shape-wise correct, uniformly dense and only incomplete, whereas airborne PCs on the other hand are not just incomplete but also have point density variations and noise.

In the past two decades, PC-based 3D building reconstruction has been extensively explored (Wang et al., 2018; Buyukdemircioglu et al., 2022), nonetheless, there are (1) not a lot of studies on non-planar (free-form) building models, (2) very few DL-based solutions. (3) Most existing generic DL-based 3D reconstruction solutions cannot work directly on airborne LiDAR building PCs since some building instances have full facade occlusions, and (4) most classical solutions are either based on a selected few building forms hence have a limited solution space or some of their generated models are geometrically erroneous and/or non-watertight due to noise or incomplete data (Nan and Wonka, 2017; Huang et al., 2022).

In this study, we propose a DL framework capable of bridging the gap given the above stated 3D building reconstruction problems. Thus, we propose the following solutions as contributions:
- A dynamic multi-scale attention-based network, which provides both local and global semantic information for complete building shape generation from partial airborne LiDAR building PCs to mitigate the occlusion, noise and sparsity problem.

- A softmax-less skip attention module to improve Point2Mesh performance by emphasizing non-local edge geometric features and/or similarities.
- A neural 3D building reconstruction framework capable of handling several building styles/complexities including free-form styles.
- We show through experimental results that the integration of PCC in a generic 3D reconstruction pipeline can bridge the gap to allow the direct reconstruction of highly occluded, sparse and/or noisy airborne LiDAR PCs into geometrically accurate surface models.

## 2. Related Work

### 2.1 Point cloud completion

The task of PCC is the process of accurately inferring the missing parts of a given incomplete PC while preserving its existing shape using geometric and semantic feature representations. Generally, most PCC models are built with input PCs that are shape-wise correct, uniformly dense and only incomplete. Popular PCC methods like FoldingNet (Yang et al., 2018), PCN (Yuan et al., 2018), AtlasNet (Groueix et al., 2018) generate complete points using the encoded features of the incomplete points' global shape. However, generic encoded global shape codes suffer from local information deficiency. Thus, Wen et al. (2020) introduced skip-attention mechanism into the classic encoder-decoder PCC pipeline to encode available local geometries – which is selectively used (based on pattern similarity) during complete point generation to boost local geometric accuracy. Similarly, Chang et al. (2021) improved the fidelity of completed points' local geometry by incorporating pointconv (Wu et al., 2018) in their completion pipeline to ensure outlying point removal and uniform distribution of generated points. Alternatively, Pan et al. (2021) approached the task of PCC via probabilistic modeling by learning global shape of PCs using distributions learned from complete PCs as a guide. Tang et al. (2022) used unsupervised learning to generate keypoints from complete points, which aided global shape preservation towards point completion and refinement. Other studies tackled PCC by using reinforcement/adversarial learning (Sarmad et al., 2019), Transformers (Zhou et al., 2022; Yan et al., 2022) or unsupervised learning (Cai et al., 2022). For effective 3D building reconstruction of airborne PCs – which unlike usual PCC model inputs, has non-uniform point density and noise – we seek to incorporate multi-scale PCC in our reconstruction framework.

### 2.2 3D Reconstruction

The task of 3D reconstruction has been tackled using various approaches and data modalities. Existing classic 3D building reconstruction approaches can loosely be broken down into two groups: geometric primitive extraction and/or assembly, and implicit fitting of polygon surfaces. The former often uses either (1) RANSAC for primitive extraction and apply geometric constraints as regularization to build primitive connections (Nan and Wonka, 2017; Coudron et al., 2018; Bauchet and Lafarge, 2020; Huang et al., 2022); (2) or employ region growing to generate super-points on which primitive can be fitted and/or assembled (Hu et al., 2018; Özkan et al., 2022; Szabo et al., 2023). RANSAC is predominantly employed over region growing because it is robust against noise, outliers, and missing data whiles region growing can be slow and sensitive to initial seeds, noise, and outliers. Even though both extraction algorithms can produce compact models, they have the demerit of being task/scene specific. Hence, they do not perform well where the set constraints are not met. The latter reconstructs 3D surfaces by implicitly fitting triangular mesh surfaces as level sets – a popular example of which is the Poisson reconstruction (Kazhdan et al., 2006). The original Poisson algorithm is computationally tasking, sensitive to sparsity, noise/outliers, and has the tendency to over-smooth the target surface thus loosing details. To mitigate these demerits, Kazhdan and Hoppe (2013) added a data fitting term to the original Poisson reconstruction objective function to explicitly interpolate points thus accounting for missing/sparse data regions consequently improving reconstructed output. Realizing that even with interpolation some spurious and/or coarse surfaces are generated due to missing data, Kazhdan et al. (2020) added a close envelope constraint to the Poisson solver to further reduce the occurrence of spurious surfaces. Most implicit approaches use volumetric data structures, but Zhao et al. (2021a) proposed a progressive discrete domain with a tetrahedron mesh instead due to its intrinsic invariance to rotations. Others also proposed parallelization (Bolitho et al., 2009) and distributed parallelization (Kazhdan and Hoppe, 2023) to manage computation complexity.

From the DL perspective, 3D reconstruction approaches can be grouped in two based on their shape output representation. That is either discretely/explicitly (e.g., voxels, points, or meshes) or continuously/implicitly (e.g. (un)signed distance functions, occupancy functions, etc.).

**Explicit methods.** Some works (Wu et al., 2015; Peng et al., 2020; Chibane et al., 2020) used voxel-based 3D CNNs for 3D reconstruction, however, it was observed that voxels consume memory cubically with increasing resolution, thus limiting grid resolutions resulting in the loss of fine shape details. As a palliative solution, other studies (Tatarchenko et al., 2017; Wang et al., 2019a) use adaptive volumetric partitioning techniques like octrees to increase resolution and manage memory complexity. Points have also been used as an alternative to represent 3D shapes (Lin et al., 2018; Choe et al., 2022a, b), but point-based representations lack the capacity to characterize topological relations, thus sometimes requiring further processing. Meshes are a common form of representation for 3D reconstruction. Hanocka et al. (2020) casted the 3D reconstruction problem as a mesh optimization problem where a mesh convex hull of an input point cloud is iteratively deformed to fit the shape of the point cloud. Gao et al. (2020) using both image and points as input also applied the mesh deformation concept on tetrahedral meshes and used point occupancy prediction to delineate object shape. Although meshes are a common form of surface representation, their reconstruction usually requires a reference template and do not respond well to missing/sparse data. As an alternative to meshes, some studies (Wang et al., 2020; Liu et al., 2021; Li et al., 2022) opted to output CAD models where each surface plane is represented by a single face. Although having slightly varying architectures, both approaches deployed 3 sub-networks to find edge/corner candidates, localize precise corner from the candidate pool and determine the correct edges connecting the corner points. Unlike other methods they both require wireframe as part of the input. Furthermore, the latter is strictly built with roof points hence a ground elevation will have to be assumed for roof boundary extrusion to obtain a full 3D building model. Kada (2022) also used building footprints deduced from boundaries of roof point sets, in addition to segmented roof faces and their slopes obtained from roof points to reconstructed 3D building models using half-space modelling. Like Li et al. (2022), a limitation of this approach is that the ground elevation of its generated models is mostly extrapolated, and the correctness of the facades are dependent on that of the roof boundary.

**Implicit approaches.** Implicit neural representations cast the 3D reconstruction problem as a binary classification problem where points in 3D space are either classified as inside or outside a continuous surface using the signed distance function (SDF) (Park et al., 2019; Hao et al., 2020; Zheng et al., 2020; Ouasfi and Boukhayma, 2022) or occupancy function (Peng et al., 2020; Erler et al., 2020; Boulch and Marlet, 2022; Williams et al., 2022) to determine a continuous decision boundary of a shape, except in the case of Atzmon et al. (2019) where particle methods are used. Owing to the global nature of DeepSDF (Park et al., 2019) representation – which limits level of detail and generalization, Chabra et al. (2020) and Erler et al. (2020) in their respective ways learned local shape priors from volumetric patches. Also, Chibane et al. (2020) argued that most signed distance and occupancy-based approaches can only model closed surfaces. Thus, the authors modelled open and closed surfaces by learning unsigned distances instead. Alternatively, sign agnostic learning (Atzmon and Lipman, 2020) and its variants (Atzmon and Lipman, 2021; Zhao et al., 2021b) also modelled both open and closed surfaces by using unsigned distance field (UDF) to learn signed representations directly from raw 3D point clouds. In 3D building reconstruction, Stucker et al. (2022) reconstructed an iso-surface of a 3D scene from a continuous occupancy field using learned embeddings of photogrammetric point cloud and ortho-photo stereo pairs. Chen et al. (2022) combined RANSAC, binary space partitioning, occupancy-based implicit DL and graph-cut in a single framework to develop a hybrid 3D building reconstruction model dubbed Point2Poly. Thus, it is evident that 3D reconstruction in DL has been studied extensively but not a lot of work has been done in relation to building modeling.

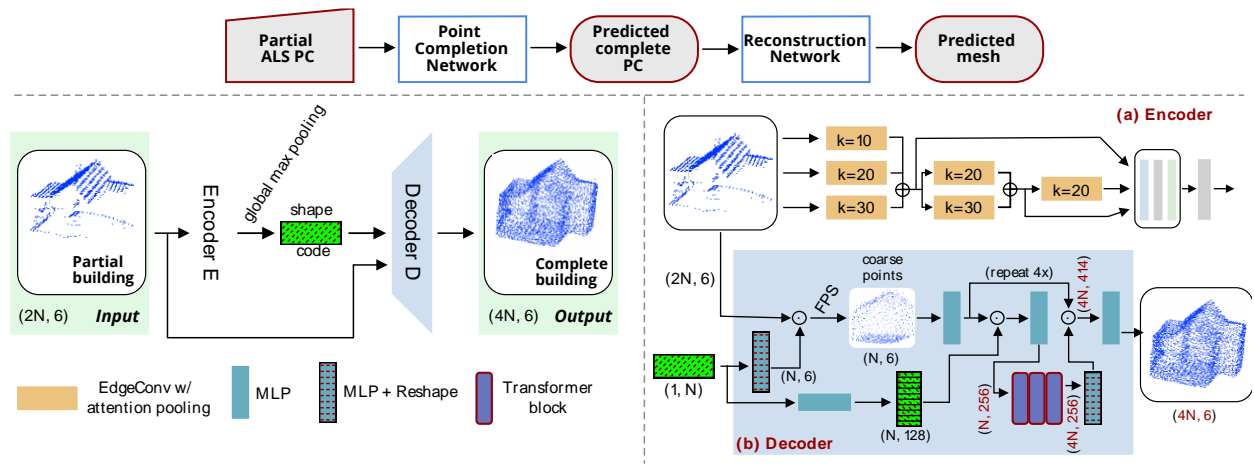

Figure 1: Proposed (top) full pipeline of APC2Mesh, (bottom left) PCC framework, (bottom right) PCC framework details.

## 3. Airborne point cloud (APC) to mesh model

Given building instances obtained from an airborne LiDAR scan – which have facade and/or roof occlusions, noise and variable point density, the goal here is to train a model to reconstruct 3D mesh models. We propose APC2Mesh, a two-stage approach which encompasses: (1) A point completion phase for filling in the areas of occlusion, mitigating noisy points, reducing sparsity and allow edge points consolidation. (2) A point modelling phase for converting the completed PCs into building mesh models. The architectural overview of the proposed framework is illustrated in Fig. 1-top.

### 3.1. APC2Mesh: Point Cloud Completion
The point completion phase of our proposed framework adopts the popular encoder-decoder network architectural design.

**Encoder.** Point local neighborhood encoding is a core part of 3D point cloud processing which ensures the embedding of local semantics into extracted feature representations. However, $k$NN queries for point neighborhood analysis are sensitive to variable point density and each query's neighborhood similarities differ in input and feature spaces. To mitigate these problems and extract locally relevant semantic cues, we design an encoder $E$ using edge convolution (Wang et al., 2019b) to extract features from local graphs at multiple scales and aggregate them based on their attention scores. Figure 1a shows our multi-scale edge convolution shape code extractor. Given an incomplete airborne building point sets $\mathcal{X} = \{x_i \mid i = 1, \ldots, n\} \in \mathbb{R}^{(n \times 6)}$ where $n$ is the total number of points and $x_i = \{x, y, z, n_x, n_y, n_z\}$ denote the coordinate and normal of a point. Let a point's neighborhood $\mathcal{N}$ at scale $s$ with $k$ points defined around query point $x_i$ be $\mathcal{N}_s^k(x_i)$. Thus, as outlined in Eqs (1) – (4), the encoder takes $\mathcal{X}$ as input and applies edge convolutions at $s$ different scales followed by an attention-based neighborhood aggregation, and an element-wise multi-scale aggregation per layer for $L$ layers. From Eqs (1) – (4), $h^\theta$ denotes a convolution block made up of convolution, normalization and activation functions. $f_s^k$ and $\alpha_s^k$ represent the point features and attention scores per neighborhood scale, respectively.

$$f_s^k = h^\theta \left( h^\theta \left( \mathcal{N}_s^k(x_i) \right) \right) \tag{1}$$

$$\alpha_s^k = \sigma \left( h^\theta (f_s^k) \right) \tag{2}$$

$$f_\alpha = \frac{1}{S} \left( \sum_{s=1}^{S} (\alpha_s^k \cdot f_s^k) \right) \tag{3}$$

$$F_g = \max \left( h^\theta \left( f_\alpha^1 \odot f_\alpha^2 \odot f_\alpha^3 \right) \right) \tag{4}$$

The attention features from the first $L - 1$ layers ($f_\alpha^1, f_\alpha^2, f_\alpha^3$) are then concatenated and projected into a higher feature dimension, followed by a global max pooling (**max**) to obtain the building shape code $F_g \in \mathbb{R}^{1024}$.

**Decoder.** Our encoder mostly focuses on learning locally attentive geometric cues. Thus, we design a decoder that can deduce both local and non-local feature correlations of building geometric cues from the extracted shape code and the input point set. Since the core concept of Transformers (Vaswani et al., 2017) is to learn the relationship of each word in a sentence to each other and then the importance of each word to the whole sentence, we adopted the use of transformers in our decoder to learn both local and global context of the extracted local geometric cues in relation to the input building point sets. Figure 1b shows the outline of our point completion decoder. The input to the decoder is the generated shape code $F_g$ and the partial point set $\mathcal{X}$ which produces completed points and their normals. Since $F_g$ is a compressed representation of all available structural information of $\mathcal{X}$, the coarse points mapped from $F_g$ tend to have a certain level of geometric degradation. To mitigate this degradation, the geometric fidelity of the input region needs to be preserve while accurately generating the missing areas. To this effect, $\mathcal{X}$ is combined with the generated initial coarse points and then a skeletal structure of better geometric fidelity is sampled using farthest point sampling (FPS) to get the coarse points. To maintain both local and global geometry, $F_g$ and the coarse points are concatenated and sent through transformer blocks to first learn local-global feature correlations and second up-sample the coarse input into a dense output. To maintain fidelity, coarse point features are combined with the up-sampled points through a skip connection. Finally, an MLP is used to project the features back to a complete $\mathcal{X}$ with an up-sample ratio of 2. The transformer block used in the decoder is defined in Eq. (5) as:

$$\begin{aligned}\mathcal{X} &= \eta\big(\mathcal{X} + \tau\,(Q, K, V)\big) \\ &= \eta\,(\mathcal{X} + \phi\,(\mathcal{X}))\end{aligned} \tag{5}$$

where $\eta(\cdot)$, $\tau(\cdot)$ and $\phi(\cdot)$ denote layer normalization, multi-head self-attention, and feed forward block all computed in a similar manner as Vaswani et al. (2017). $Q = XW_Q$, $K = XW_K$, and $V = XW_V$, where $Q, K, V$ represent the query, key and value matrices, respectively. $W_Q \in \mathbb{R}^{c_q \times c_q}$, $W_K \in \mathbb{R}^{c_k \times c_k}$ and $W_V \in \mathbb{R}^{c_v \times c_v}$ are linear transformation matrices. $c_q$, $c_k$ and $c_v$ represent query, key and value matrix dimensions, respectively.

**Loss.** Since our point completion estimates both points and normals, we defined a compound loss ($\mathcal{L}_{comp}$) made up of the sum of coarse ($\mathcal{L}_{CD}^c$) and fine ($\mathcal{L}_{CD}^f$) chamfer distances as point loss and a squared $L_2$-norm as normal loss ($\mathcal{L}^n$) in Eq. (6) as:

$$\mathcal{L}_{comp} = \mathcal{L}_{CD}^c + \mathcal{L}_{CD}^f + \mathcal{L}^n \tag{6}$$

where $\mathcal{L}_{CD}^c$ and $\mathcal{L}_{CD}^f$ denote chamfer distances for coarse ($S_{coarse}$) and fine ($S_{fine}$) completed points, respectively. Thus, chamfer distance (CD), $\mathcal{L}_{CD}^c$ and $\mathcal{L}_{CD}^f$ are defined as:

$$CD(S_1, S_2) = \frac{1}{|S_1|} \sum_{\hat{y} \in S_1} \min_{y \in S_2} \|\hat{y} - y\|_2 + \frac{1}{|S_2|} \sum_{y \in S_2} \min_{\hat{y} \in S_1} \|\hat{y} - y\|_2 \tag{7}$$

$$\mathcal{L}_{CD}^c = CD(S_{coarse}, S_{GT}) \tag{8}$$

$$\mathcal{L}_{CD}^f = CD(S_{fine}, S_{GT}) \tag{9}$$

where $S_2$ represents the groundtruth ($S_{GT}$), and $S_1$ represents either $S_{coarse}$ or $S_{fine}$.
The normal loss ($\mathcal{L}^n$) in Eq. (6) is defined as:

$$\mathcal{L}^n = \frac{1}{N} \sum_{\hat{y}^n \in P^n, y^n \in P_{GT}^n} \|\hat{y}^n - y^n\|_2^2 \tag{10}$$

where $P_{GT}^n$, $P^n$ and $N$ denote groundtruth (GT) point normals, predicted point normals and point size, respectively.

### 3.2. APC2Mesh: Building Shape Reconstruction

The reconstruction phase of our framework casts the 3D building reconstruction as a mesh deformation problem where convolutional kernels are learnt directly on mesh edges to obtain non-uniform, local geodesic features invariant to scale, rotation and translation – which is used in computing edge/vertex position shifts. However, different from Hanocka et al. (2020), we employ linearized skip attention mechanism to emphasize geometric similarities and edge features.

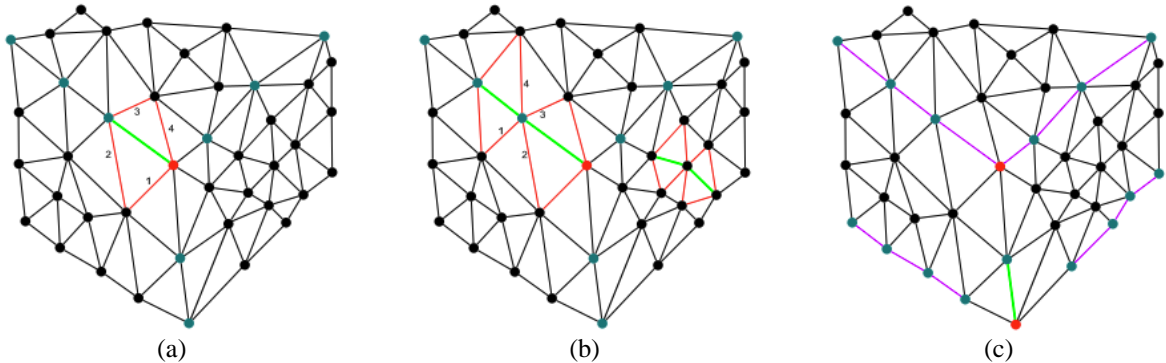

Figure 2: A sample mesh illustrating (a) the 1-ring neighborhood employed in MeshConv; (b) the limited receptive field characteristic of MeshConv; and (c) the absence of non-local connectivity among MeshConv extracted features. The color scheme is as follows: green lines represent query edges, red lines denote 1-ring neighbors, purple lines signify significant non-local edges, red points indicate shape corners, and green points represent shape edges.

The vanilla Point2Mesh lacks non-local edge interdependency as it uses mesh convolution (MeshConv) from MeshCNN (Hanocka et al. 2019) to learn local edge descriptors. For a query edge (Fig. 2(a) green line) in a mesh, MeshConv is achieved by performing convolutions on the query edge and it's 1-ring neighbourhood (Fig. 2(a) red lines). Consequently, the receptive field of MeshConv is very small, hence, limiting the interconnectivity among the extracted edge features. As illustrated in Fig. 2(b), even the relationship between the extracted features of two neighbouring green edges is not direct; rather, it is indirectly linked through the features of edges 1, 2 and edges 3, 4.

Thus, the question arises: *How can we establish long-range dependencies among all edges to effectively underscore important mesh properties like shape edges* (Fig. 2(c))*?* Our proposed solution is the integration of skip self-attention into Point2Mesh. Considering the classic dot-product self-attention exhibits a quadratic memory complexity with respect to the number of edges, we opted to define an $L_2$ normalization attention mechanism, which offers an almost linear computation complexity. The complete workflow of our 3D building reconstruction step is as illustrated in Fig. 3 and described in detail below.

A mesh convex hull is first created from the output of our PCC model to serve as an initial mesh. Based on the connectivity of this initial mesh, random coordinates representing the edge $e$ (vertex pairs) of the initial mesh is generated to serve as added input to the network. Based on the mesh and edges, mesh convolutions followed by normalization and activation functions are applied over $L$ layers to produce edge features. The UNet structure is adopted in this study, however, to accentuate and effectively communicate features of importance from the encoder with the decoder, we introduce a skip self-attention (SSA) module into our framework along the encoder-decoder skip connections. The SSA not only accentuates both local and non-local self-similarity, but also prioritizes through attention scores edge features – which later translates into a more efficient mesh edge/vertex shift and distribution thus allowing for more accurate reconstructed geometry.

Given that the classic dot product attention (Vaswani et al., 2017) has a quadratic memory complexity and multiplication is associative, we compute a softmax-less linear attention in Eq. (11) as:

$$L_{att}^i = \frac{1}{N_e} \|Q\|_2 \left( \|K^T\|_2 \cdot V \right) \qquad (11)$$

Given an encoder layer $L^i$, we compute $K$, $Q$ and $V$ matrices of the extracted edge features $L_{ef}^i \in \mathbb{R}^{N_e \times c}$ as $Q = f_{mc}(L_{ef}^i W_Q)$, $K = f_{mc}(L_{ef}^i W_K)$, and $V = f_{mc}(L_{ef}^i W_V)$. $N_e$ denotes the number of mesh edges and $f_{mc}$ is mesh convolution with normalization and activation. $W_Q \in \mathbb{R}^{c \times c_q}$, $W_K \in \mathbb{R}^{c \times c_k}$ and $W_V \in \mathbb{R}^{c \times c_v}$ are transformation matrices with $c_k$, $c_q$ and $c_v$ representing $K$, $Q$ and $V$ matrix dimensions, respectively.

The output per layer of each SSA module is layer-by-layer fused with progressively up-sampled output of the last encoder layer to obtain the final decoded edges (vertex pairs) $\hat{e}$. These edges $\hat{e}$ represent $e$ in displaced positions. With the connectivity of the initial mesh already established; the individual vertex shifts required to move the initial mesh towards the point cloud can be computed as:

$$\Delta v_i = v_i^0 - \hat{v}_i, \quad \text{s.t.} \quad \hat{v}_i = \frac{1}{\mathcal{N}} \sum_{j \in \mathcal{N}} \hat{e}_j(v_i) \qquad (12)$$

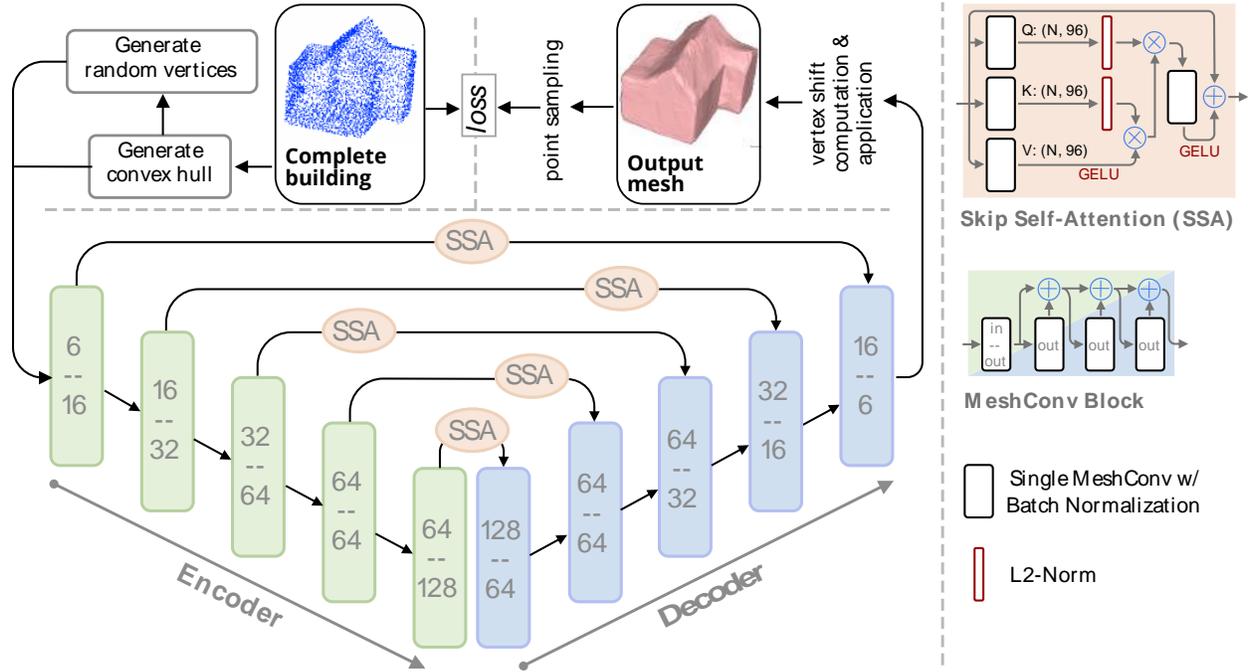

Figure 3: Graphical illustrations of the APC2Mesh 3D building shape reconstruction framework

where $v_i^0$ and $\hat{v}_i$ denote the initial mesh vertex and shifted mesh vertex, respectively. $\mathcal{N}$ is the number of edges incident from $v_i^0$, and $\hat{e}_j(v_i)$ is the vertex position of $v_i^0$ along $\hat{e}_j$. Thus, the encoder-decoder network can be viewed as a weight parameterization process between the initial convex hull mesh and the complete point cloud.

**Loss.** The training of the reconstruction framework is driven by point chamfer distance $\mathcal{L}_{CD}^p$, normal cosine similarity $\mathcal{L}^d$, and local uniformity ($\mathcal{L}^u$) losses yielding a total loss defined in Eq. (13) as:

$$\mathcal{L}_{total} = \mathcal{L}_{CD}^p + \mathcal{L}^d + \mathcal{L}^u \tag{13}$$

where $\mathcal{L}_{CD}^p$ shares the same definition as Eq. (7). $\mathcal{L}^d$ and $\mathcal{L}^u$ are defined in Eq. (14) and Eq. (15), respectively as:

$$\mathcal{L}^d = -1 \cdot \left| \frac{P^n \cdot P_{GT}^n}{\|P^n\|_2 \cdot \|P_{GT}^n\|_2} \right| \cdot \lambda_1 \tag{14}$$

$$\mathcal{L}^u = \left( \frac{1}{N_f} \left\| A_f - A_f^{en} \right\|_1 \right) \cdot \lambda_2 \tag{15}$$

where $\lambda_1$ and $\lambda_2$ are weight terms set to 0.1. $N_f$ denotes the number of mesh faces, while $A_f$ and $A_f^{en}$ denote the areas of a mesh face and its edge neighbor, respectively.

## 4. Data, Experiments and Results

In this section, we present experimental evaluations to show the efficacy of our proposed framework. This includes details of the datasets used, implementation details of APC2Mesh, experimental results, and ablation studies. In our evaluations, we mainly use Chamfer distance and/or root mean squared error (RMSE) as metrics to compare predictions to ground truths. Sections 4.3 and 4.4 focus on PCC and 3D building reconstruction evaluations, respectively.

### 4.1. Experimental Data

High quality building point clouds and their corresponding 3D surface models are needed for the training and evaluation of DL-based 3D reconstruction. Thus, we used data from the Building3D dataset (Wang et al., 2023) for training and evaluation. The Building3D dataset encompasses airborne LiDAR point clouds (APCs) of building instances and their corresponding mesh models covering the republic of Estonia. For this study, we limited our training and evaluation to 6 cities, namely Loksa, Sillamae, Paide, Tartu, Hiiumaa and Haapsalu. For detailed dataset description, statistics, and extra visuals; please refer to the source (i.e., Wang et al. (2023), and the building3D dataset website (https://building3d.ucalgary.ca/index.php)).

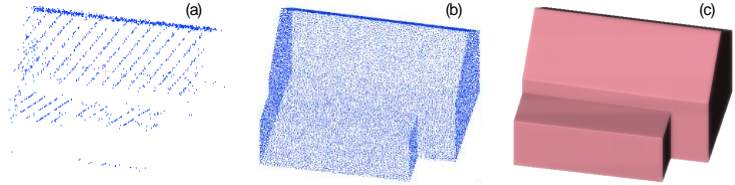

Figure 4: A sample of our experimental data. Input APC (a), GT point set (b) and GT mesh model (c).

Some building instances in the Building3D dataset have noisy points, so to eliminate them, we used point-to-mesh distance thresholding. Specifically, we first normalized point set instances and their corresponding mesh models into a unit sphere. Second, since the point-mesh instances are in the same coordinate reference frame and aligned, we up- and down-scale the meshes by a fixed threshold. Any point that falls within the down-scaled mesh or outside the up-scaled mesh is considered a noisy point and removed. Furthermore, the APC building instances either have more or less points than our chosen input size of 2048, so we down- or up-sampled, respectively, each instance to a fixed size of 2048. To generate the groundtruth (GT) point set instances for training the PCC network, we sampled 16384 points per instance from the mesh models. These GT point instances and their corresponding normals are also used in positional RMSE (pRMSE) and directional RMSE (dRMSE) computations.

Within the 6 cities chosen as our study area, we selected 10384 and 404 sample pairs of building instances for PCC training and PCC testing/3D reconstruction, respectively. The training instances were taken from three cities (i.e., Sillamae, Paide and Haapsalu) and the testing set was selected from three different environments with varying building styles and levels of complexity. That is, an island (Hiiumaa), a beach city (Loksa) and the less developed part of a land-locked city (Tartu). Figure 4 shows an example of the input APC and its corresponding point and mesh GT.

## 4.2. Implementation Details

All model training and evaluation processes were conducted on an 12GB Titan X GPU. Our point completion network takes incomplete point sets with $8 \times 2048$ points, where 8 is the batch size, process them at 3 different scales (10, 20, 30) and output complete point sets of size $8 \times 4096$. We trained the PCC model for 145 epochs using an initial learning rate (lr) of 0.0006 in a cosine annealing lr scheduler with warmup. In our reconstruction step, we train each instance for 1000 iterations to obtain our output mesh model. For a balance between computation time and surface detail/resolution, we performed deformation at two levels of mesh resolution. Specifically, we set the number of faces of our input convex hull mesh to 3000 for the first 500 iterations during which the initial random vertices are maintained. Before the second 500 iterations, we up-sample the deformed mesh thus far to 4000 faces, compute new random vertices and repeat the entire process using the up-sampled mesh and its corresponding random vertices for the second 500 iterations. To compute network loss per iteration during training, we sample 4096 points from the deformed mesh and gave it as input, along with the corresponding completed points, to Eq. (13). Final evaluation metrics are computed between points sampled from the reconstructed mesh and the GT mesh.

## 4.3. PCC Evaluation

Table 1 outlines the results of our ablation study on the PCC model's components in relation to completion performance. The bottom row in Table 1 represents the obtained result of the proposed APC2Mesh PCC model. As can be seen, the resampling of concatenated input and predicted coarse points using farthest point sampling (FPS) aided in geometric fidelity preservation. Additionally, the inclusion of point normal estimation in point completion training also benefited the geometric fidelity of predicted complete point sets. For our single scale experiment, a local neighborhood ($k$) of 20 was used. From Table 1, the use of multi-scaling has about the same magnitude of impact on completion accuracy as attention pooling. The selection of scale "s" and the number of neighbors "k" was determined empirically, considering the experimental values presented in Table 2. Figures 5, 6 and 7 show the qualitative PCC results of the three different testing regions with varying levels of building complexity. As can be seen, our PCC model is able to complete the input APCs – of varying complexities and styles – with high geometric fidelity.

Table 1: Effects of the individual components of our PCC model on completion results. **I-CC&R:** input–coarse points concatenation and resampling; **AP:** Attention pooling, if unchecked means max pooling was used; **N:** Point normal estimation included in training; **MS:** multi-scaling. The smaller the $\mathcal{L}_{CD}^{f}$ value the better.

| I-CC&R | AP | N | MS | $\mathcal{L}_{CD}^{f}$ ($\times 10^{-3}$) |
|---|---|---|---|---|
| | ✓ | ✓ | | 0.83 |
| | ✓ | ✓ | ✓ | 2.99 |
| ✓ | ✓ | | ✓ | 1.27 |
| ✓ | | ✓ | ✓ | 0.82 |
| ✓ | ✓ | ✓ | ✓ | 0.78 |

Table 2: Ablation studies on the choice of scale $s$ and the number of neighbors $k$ in our PCC model. $s = \text{size}(k)$ and $\mathcal{L}_{CD}^{f}$ values are ($\times 10^{-3}$)

| $k$ | [20] | [16, 24] | [8, 16, 24] | [10, 20, 30] |
|---|---|---|---|---|
| $\mathcal{L}_{CD}^{f}$ | 0.83 | 0.81 | 0.80 | **0.78** |

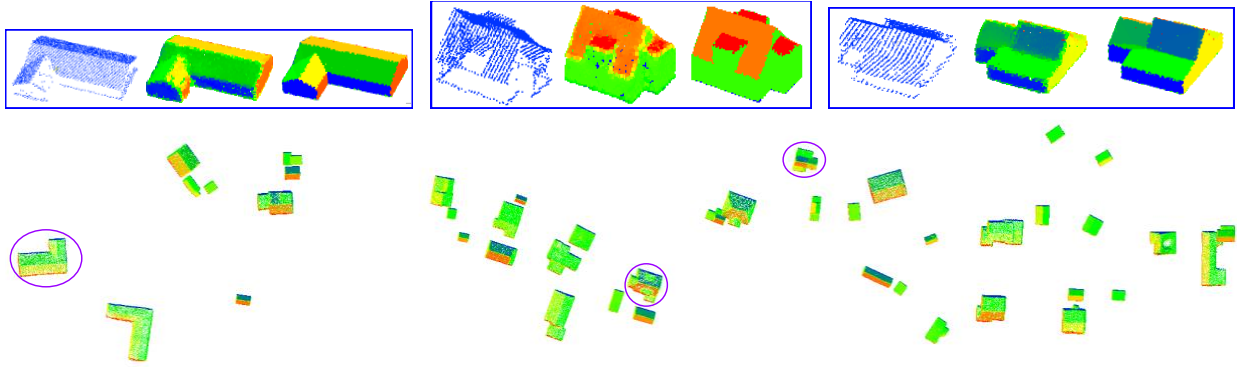

Figure 5: Qualitative results of APC2Mesh PCC testing scene 1 (an island). Each zoomed image inset represent (from left to right) partial APC input, predicted completed output, and GT. The random colours displayed in the predicted and GT point are used to highlight the point normals.

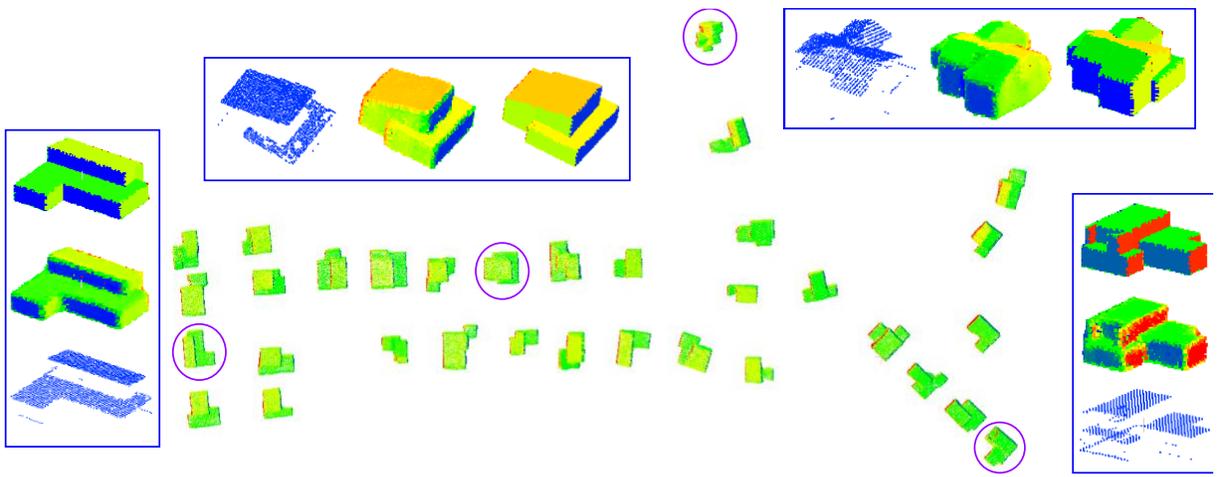

Figure 6: Qualitative results of APC2Mesh PCC testing scene 2 (beach city). Each zoomed image inset represent (from left to right or down to top) partial APC input, predicted completed output, and GT. The random colours displayed in the predicted and GT point are used to highlight the point normals.

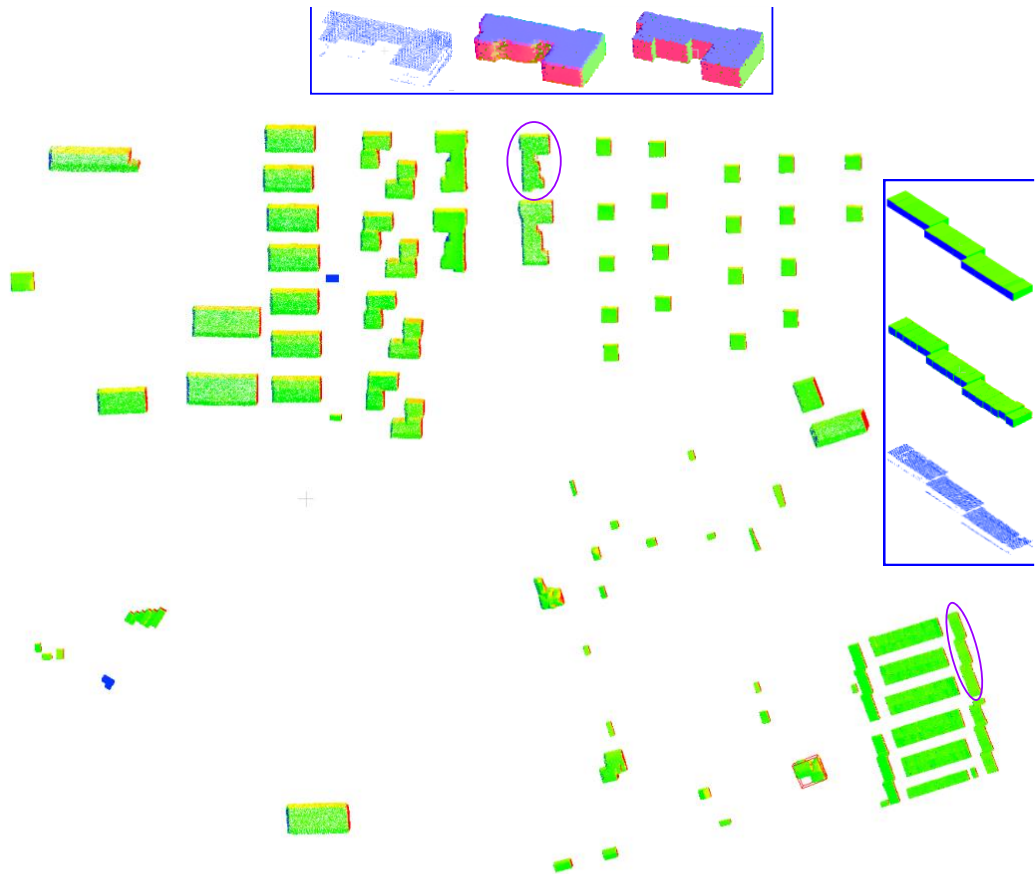

Figure 7: Qualitative results of APC2Mesh PCC testing scene 3 (land-locked city – less developed area). Each zoomed image inset represent (from left to right or down to top) partial APC input, predicted completed output, and GT. The random colors displayed in the predicted and GT point are used to highlight the point normals.

We compare our point completion method to other completion methods in the 3D building reconstruction domain, specifically the recent work of Zhao et al. (2023) which leveraged the planarity and regularity geometric properties of buildings as regularization constraints for point completion. It should be noted that although airborne, Zhao's method was built on UAV-based point cloud data (flying height: 120m; point density: 273 points/m$^2$) and our experimental data was captured using an airplane at a flying height of 2600m and has a point density of ~30 points/m$^2$. Figure 8 show some of the results obtained.

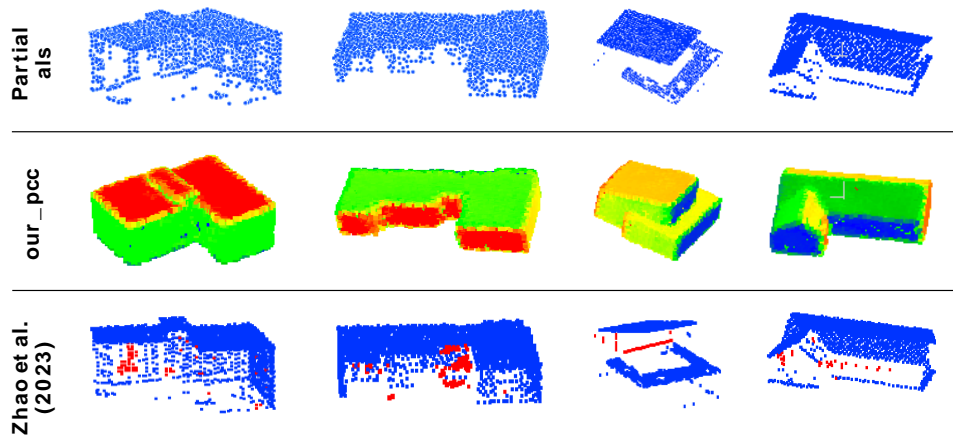

Figure 8: Comparison of our PCC results to Zhao et al. (2023). Bottom row: APC (blue) + Zhao's completion (red); Middle row: our PCC outputs colored for better visualization of planes.

According to Zhao et al. (2023), the cloth simulation algorithm used to patch the point cloud occlusions requires the input point cloud to have a specific orientation, achievable through an attitude adjustment procedure. Furthermore, the employed attitude adjustment method(s) require the presence of at least one clean plane in the x- or y-axis to work with – which the UAV data would have provided, given its point density. The deficiency of this method becomes evident in the scenario where there are little to no discernible facade points to compute an accurate plane from – a scenario in which majority of our experimental data fall.

Given that an airplane usually captures point cloud with so little façade information, and roof/ground edge points that can serve as façade information are fuzzy with no clean edge lines, Zhao's method generated spurious facades in addition to correct ones. Thus, in the case of high sparsity, noise, and severe façade occlusions, Zhao's method will yield sub-optimal results. Compared to Zhao's our PCC method yielded better results because it is learning-based and not constrained by geometric rules.

### *4.4. 3D Building Reconstruction Evaluation*

To determine the efficacy of our proposed framework, we compute the positional and directional (normals) root mean square error (i.e., pRMSE and dRMSE) between the ground truth and reconstructed meshes. We provide both (qualitative) individual building instance errors (in the form of heat maps) and average quantitative errors. Based on these metrics, we compared our experimental results to some existing methods (Fig. 9 & Table 3). All experiments in this section excluding City3D were conducted with completed points from our PCC model as input. City3D experiments – as a traditional approach baseline – were conducted using the arbitrary partial airborne point sets.

**Improved Point2Mesh (P2M).** We compared our improved P2M (Ours) with the vanilla P2M (P2M) (Hanocka et al., 2020) – whose core deformation concept we adopted and improved for 3D building reconstruction – to evaluate the impact of our modifications. From Fig. 9, APC2Mesh produced building models with a slightly lower positional and directional error bounds for most building instances. Meaning the position and orientation of mesh vertices and faces are relatively closer to the groundtruth compared to those of P2M. The evidence of which can be seen in the reconstructed meshes where APC2Mesh has slightly less smooth corners and edges compared to P2M. Thus, the introduced SSA module helps maximize mesh vertex position and face direction accuracy in planar surfaces and edge/corner features thus allowing for more accurate reconstructed buildings. It also boosts mesh vertex distribution efficiency to space out in regions of high feature similarities since less surface detail is required, and cluster up at regions of high feature variance to better depict local surface shape. Although relatively marginal, our improvements to the vanilla P2M yielded a 0.002m and 0.06 decrease in positional and directional errors, respectively.

We also investigated the method of normal estimation, as the deformation of meshes as implemented in APC2Mesh requires the use of normals. Particularly, we examined the scenario where the PCC points were accompanied by traditionally computed normals instead of the PCC estimated ones (hereon referred to as deep normals) for 3D reconstruction. As outlined in Table 3, employing deep normals led to markedly improved accuracy. Notably, we observed an approximate decrease of $\pm 0.055$ in positional errors for both Ours* and P2M*. Additionally, it yielded an approximate decrease of $\pm 1.386$ in directional errors for both Ours* and P2M*.

**Comparison with existing methods.** Compared to our improved P2M or its vanilla variant, point2surf (P2S) (Erler et al., 2020) – a patch-based SDF approach – produced mesh models with relatively high number of faces (i.e., ~80000 on average) yet it has less smooth plane surfaces as shown by the numerous green colored specks on its directional RMSE image (Fig. 9). Additionally, the high-resolution output of P2S models make them relatively heavier in disk storage size (i.e., ~1.8MB compared to our ~123KB). Although points2poly (P2P) (Chen et al., 2022) and City3D produced compact models with very few faces, their reconstructed models' correctness is in big part dependent on input points' density, boundary accuracy and extracted primitive accuracy as evidenced by the erroneous local geometry and face normal orientations (red circle highlight) observed in P2P and City3D output meshes. Unlike the other approaches, the neural unsigned distance field (NDF) outputs a very dense PC (~1 mil.) which is post-processed to generate very high-resolution mesh models (~100K faces) with well-defined edges and corners. Nonetheless, the very high resolution that accentuates edge/corners also results in a very heavy mesh which occupies a significant amount of disk storage (~16.5MB).

In addition to the qualitative results in Fig. 9, we also provide quantitative results in Table 3 – which shows that our improved P2M (Ours) outperforms all the other methods in directional RMSE and ranked second in terms of positional RMSE. Compared to the other methods, mesh deformation can be time consuming given its iterative nature. Nonetheless, that does not diminish its performance capacity.

Table 3: Quantitative comparison of APC2Mesh 3D building reconstruction results with other existing methods (x10). Ours*: APC2Mesh without PCC normal. P2M*: vanilla P2M without PCC normal. Time: inference reconstruction time in minutes.

|  | **Ours** | **Ours*** | **P2M** | **P2M*** | **P2S** | **P2P** | **NDF** | **City3D** |
|---|---|---|---|---|---|---|---|---|
| pRMSE | 0.134 | 0.191 | 0.136 | 0.194 | 0.140 | 0.295 | 0.125 | 0.901 |
| dRMSE | 1.581 | 2.977 | 1.641 | 3.017 | 2.619 | 10.77 | 1.784 | 8.431 |
| Time (mins) | 10.7 | 10.9 | 6.59 | 6.74 | 0.62 | 0.65 | 0.81 | 1.07 |

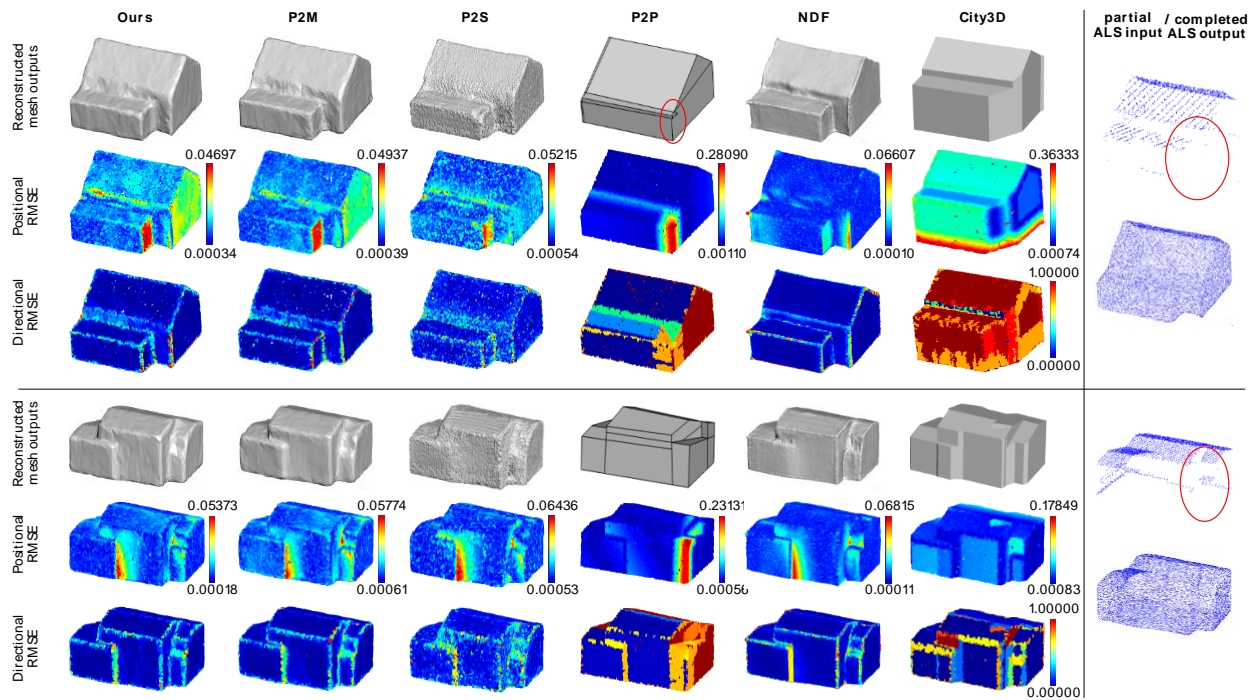

Figure 9: Qualitative comparison of APC2Mesh 3D building reconstruction results and performance errors to other existing methods. Red circular highlights indicate areas of roof occlusions in building instance. Red circle highlight indicates an area of significant local geometric error in reconstructed building instance. dRMSE error map/legend is normalized.

**APC2Mesh scene-level performance.** We further evaluated the performance of APC2Mesh at the scene-level on the three PCC testing scenes (Figs. 5, 6, 7). Selected with varying difficulty/complexity in mind, scene 1 is an island with mixed variety of buildings, scene 2 is a beach city with modern complex residential buildings and scene 3 is a less developed land-locked town with relatively simple buildings. Table 4 shows the quantitative results obtained from each scene. Seeing the minimal difference in results of the 3 scenes despite their varying level of complexity, we can say that the APC2Mesh is capable of reconstructing APC building instances with varying degrees of occlusion, complexities, and/or styles with minimal errors in geometric fidelity.

Table 4: APC2Mesh 3D building reconstruction results at different test scenes.

|  | Scene 1 | Scene 2 | Scene 3 |
|---|---|---|---|
| pRMSE | 0.0154 | 0.0196 | 0.0130 |
| dRMSE | 0.1371 | 0.2329 | 0.1015 |

**3D building reconstruction without PCC integration.** To substantiate the necessity of PCC integration in a 3D building reconstruction pipeline, we analyzed the reconstruction performance of APC2Mesh without the point completion step. Additionally, we investigated the impact of PCC coverage on 3D building reconstruction by randomly removing various percentages of PCC-filled points from previously occluded regions and then performing 3D building reconstruction. The results presented in Table 5 highlights the advantages of incorporating PCC in the 3D building reconstruction pipeline – which manifested as a reduction in positional and directional reconstruction errors by 0.031 and 1.021, respectively.

Table 5: Quantitative APC2Mesh reconstruction results on the impact of PCC integration and coverage (×10).

|       | Ours  | 75%   | 50%   | 25%   | APC (0%) |
|-------|-------|-------|-------|-------|----------|
| pRMSE | 0.134 | 0.162 | 0.158 | 0.200 | 0.442    |
| dRMSE | 1.581 | 3.082 | 3.554 | 5.802 | 11.789   |

Moreover, we demonstrate in Fig. 10 that APC2Mesh without PCC leads to incomplete mesh models, particularly in areas with substantial occlusion. This deficiency becomes more pronounced when input APCs have significant occlusions, such as missing roof parts, floors, and multiple facades (e.g., Fig. 9-right & 10). It is worth noting that the neural SDF methods (i.e., P2S and P2P) - which require closed surfaces to function effectively – will most likely produce suboptimal inference without PCC integration as made evident in Fig. 11. Similarly, the UDF and deformation methods will leave undesirable holes in the output mesh. Thus, these observations and results underscore the necessity of PCC integration in a neural 3D reconstruction pipeline for building reconstruction from APCs.

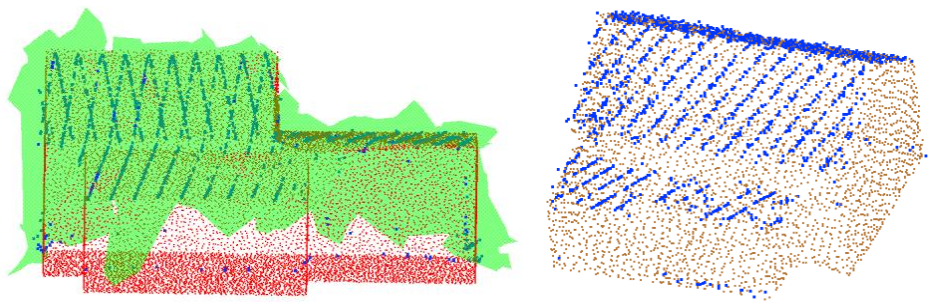

Figure 10: (left) APC2Mesh 3D reconstruction results without PCC integration. APC (blue), GT point set (red), and final reconstructed mesh (deformed convex hull (green)). Convex hull gets torn in places of significant occlusion (e.g.: roof parts, floor, facades). (right) A case of significant APC occlusion (blue) with corresponding PCC result (brown).

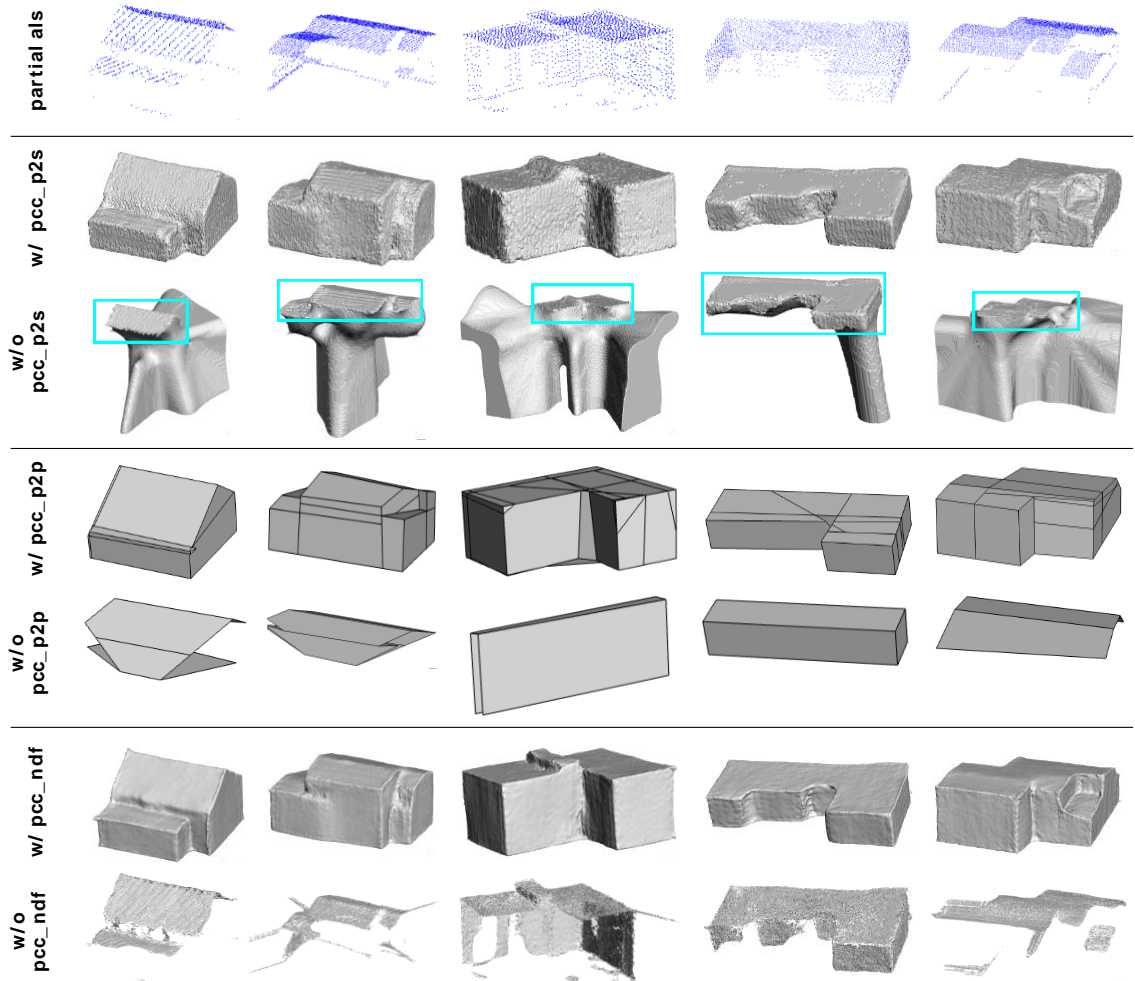

Figure 11: 3D reconstruction results with (w/) and without (w/o) PCC integration for P2S, P2P and NDF methods. The cyan colored rectangle areas highlight the reconstructed roof shapes amidst the malformations.

From Fig. 11, it is evident that the little to no availability of façade points in the input APC (w/o PCC integration) led to poor results for P2S, P2P and NDF. Primarily, these methods were only able to reconstruct the roof surfaces, albeit with significant artifacts or malformations at roof edges and façade areas, particularly in the case of P2S and NDF. Since P2P depends on RANSAC planes and the output point cloud from P2S for its final reconstruction, little to on availability of façade information from both sources resulted in its poor outcomes (w/o pcc_p2p). Since an unsigned distance field approach like NDF is capable of processing open meshes, it stands to reason that if given a partial APC input, it will perceive it as a shape with open surfaces and reconstruct it as such. Thus, the integration of a PCC step is prudent.

**APC2Mesh performance on complex/tall building instances.** To see the performance of APC2Mesh in the context of complex and/or tall building instances, we conducted a separate evaluation specifically on a selected number of complex/tall building instances from the 3 test cities/environments mentioned in section 4.1. Table 6 shows the obtained results which reveals a slight decrease in model performance with respect to complex/tall. Notably, we observed an overall 0.018 m and 0.843 increase in pRMSE and dRMSE magnitudes, respectively, compared to the primary test set. Furthermore, the complex/tall results exhibit a marginally higher minimum bound for both pRMSE and dRMSE compared to the primary test set. This rise in the minimum bound can be rationalized by the inherent geometrical intricacies prevalent in the complex/tall set, comprising predominantly of challenging building instances. The qualitative results from the complex/tall evaluation are as displayed in Fig. 12. From the visual representation, we found that the edges/corners of some of the complex building instances are a little more pronounced relative to their simpler counterparts. This subtle distinction is evidenced by the slightly lower value of complex/tall test set's pRMSE max value compared to that of the primary test set.

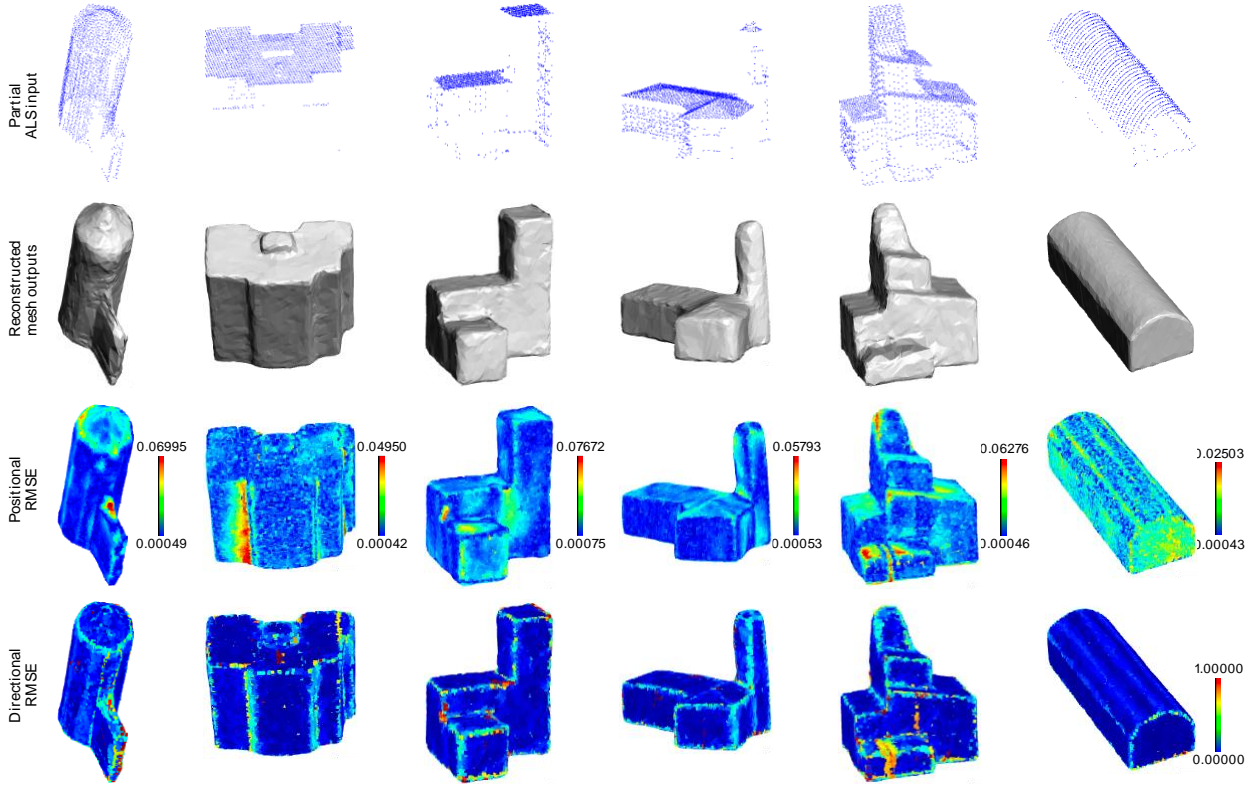

Figure 12: Qualitative comparison of APC2Mesh 3D building reconstruction results and performance errors on selected complex or tall building instances. dRMSE error map/legend is normalized.

Table 6: Quantitative results of APC2Mesh when evaluated on complex/tall buildings (x10)

|  | Primary (avg) | Primary (min/max) | Complex/tall (avg) | Complex/tall (min/max) |
|---|---|---|---|---|
| pRMSE (m) | 0.134 | 0.00067 / 4.960 | 0.152 | 0.00097/4.353 |
| dRMSE | 1.581 | 0.00010/20.00 | 2.423 | 0.00109/20.00 |

## 4.5. Ablation Studies

**Effect of affinity matrix size on attention in 3D building reconstruction.** Given that we employed a more compact representation of self-attention, we study the effects of the affinity ($KV$) matrix size ($c_k$; $c_q$) on 3D building reconstruction performance. We set $c_k$; $c_q$ to a fixed value across all layers per experiment. Table 7 shows the results obtained over different $c_k$; $c_q$ values. As expected, error metrics reduce with increasing matrix size, however it tappers off and increases between size 96 and 128 – indicating that self-attention affinity has a saturation point in relation to model performance. We also considered the time taken to reconstruct each building instance given the various affinity matrix sizes. It was observed that the affinity matrix size also affects reconstruction time – i.e., the larger the affinity matrix, the longer the reconstruction time.

**Effect of attention pooling in PCC.** We also looked at the performance of our PCC model in relation to the pooling strategy used. Max and attention pooling attained chamfer distances of $8.2 \times 10^{-4}$ and $7.8 \times 10^{-4}$, respectively. Thus, by swapping out max with attention pooling we attained a $0.4 \times 10^{-4}$ decrease in error.

Table 7: Effects of affinity matrix dimension on 3D building reconstruction performance ($\times 10$)

| Metric | Matrix sizes | | | |
|---|---|---|---|---|
|  | 32 | 64 | 96 | 128 |
| pRMSE | 0.1348 | 0.1346 | 0.1340 | 0.1345 |
| dRMSE | 1.595 | 1.588 | 1.581 | 1.587 |
| Time (mins) | 8.4 | 8.5 | 10.7 | 11.8 |

## 5. Conclusions and Discussion

The inherent flexibility of meshes and the non-parametric nature of our approach indicates APC2Mesh's potential to generalize to diverse building styles and geometries. Notably, our approach operates without the need for extra variable definitions like plane primitives (P2P) and signed distances (P2S).

Furthermore, our proposed integration of PCC not only bridges the gap between neural point completion and surface reconstruction, but also addresses the challenges posed by occlusion, noise and variable point density in generating building models directly from ALS point sets. It also stands as a strong candidate for single-big projects such as the reconstruction of historical buildings.

Despite the merits of our proposed approach, some of APC2Mesh's outputs are not immune to minimal edge/corner smoothening, especially where complex/tall building instances are concerned. Furthermore, a fixed (low) number of input points can count as a limitation of APC2Mesh – which is improvable given a large enough computation resource.

Recognizing the promising prospects of our proposed APC2Mesh, our future work will focus on addressing the limitations outlined, and refining APC 3D building reconstruction performance – especially corners/edges. Overall, we believe the versatility and effectiveness of our approach will open new avenues for the use of deep learning in diverse building modeling applications.

## Acknowledgements

This work was supported by the National Natural Science Foundation of China under Grant 42071443.

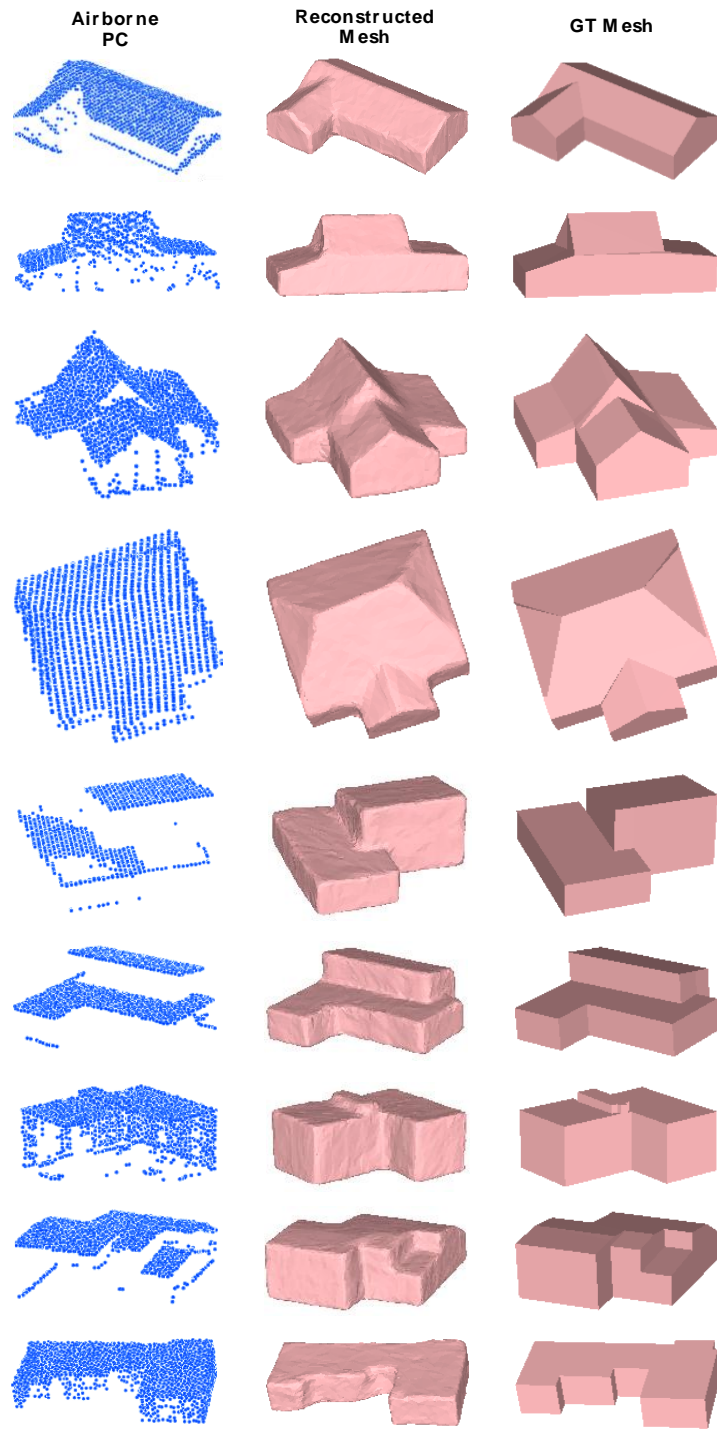

Figure 13: Some more qualitative 3D building reconstruction results of the APC2Mesh framework.